\documentclass[letterpaper, 10 pt, conference]{ieeeconf}  %

\IEEEoverridecommandlockouts                              %

\usepackage{times}
\usepackage{latexsym}
\usepackage{helvet} %
\usepackage{courier}  %
\usepackage[hyphens]{url}  %
\usepackage{multirow}
\usepackage{multicol}
\usepackage{graphicx}  %
\usepackage{url}
\usepackage{CJK}
\usepackage{float}
\usepackage{indentfirst}
\usepackage{booktabs}
\usepackage{caption}
\usepackage[T1]{fontenc}
\usepackage[utf8]{inputenc}
\usepackage{authblk}
\usepackage{capt-of,etoolbox}
\usepackage[labelsep=period]{caption}
\usepackage{amssymb}

\usepackage{amsmath,amsfonts,bm}

\def\eqref#1{equation~\ref{#1}}

\def\1{\bm{1}}

\DeclareMathAlphabet{\mathsfit}{\encodingdefault}{\sfdefault}{m}{sl}
\SetMathAlphabet{\mathsfit}{bold}{\encodingdefault}{\sfdefault}{bx}{n}

\usepackage[linkcolor=cyan,citecolor=green,]{hyperref}
\usepackage{hyperref}
\usepackage{url}
\usepackage{graphicx}
\usepackage{multirow}
\usepackage{multicol}
\usepackage{subfigure}
\usepackage{wrapfig}
\usepackage{float}

\title{\LARGE \bf
PSDet: Efficient and Universal Parking Slot Detection
}
\author{Zizhang Wu\textsuperscript{1} \qquad Weiwei Sun\textsuperscript{2} \qquad Man Wang\textsuperscript{1} \qquad Xiaoquan Wang\textsuperscript{1} \qquad Lizhu Ding \textsuperscript{1} \qquad Fan Wang\textsuperscript{1}  
\\
{\small \textsuperscript{1}Zongmu Technology \qquad \textsuperscript{2}University of Victoria}
\\
{\tt \small \{zizhang.wu, man.wang, xiaoquan.wang, lizhu.ding, fan.wang\}@zongmutech.com, \{weiweisun\}@uvic.ca}

}

\begin{document}

\maketitle
\thispagestyle{empty}
\pagestyle{empty}

\begin{abstract}
While real-time parking slot detection plays a critical role in valet parking systems, existing methods have limited success in real-world application. We argue two reasons accounting for the unsatisfactory performance: \romannumeral1, The available datasets have limited diversity, which causes the low generalization ability. \romannumeral2, Expert knowledge for parking slot detection is under-estimated.
Thus, we annotate a large-scale benchmark for training the network and release it for the benefit of community. Driven by the observation of various parking lots in our benchmark, we propose the circular descriptor to regress the coordinates of parking slot vertexes and accordingly localize slots accurately. 
To further boost the performance, we develop a two-stage deep architecture to localize vertexes in the coarse-to-fine manner. In our benchmark and other datasets, it achieves the state-of-the-art accuracy while being real-time in practice. Benchmark is available at: \href{https://github.com/wuzzh/Parking-slot-dataset}{https://github.com/wuzzh/Parking-slot-dataset}

\end{abstract}
\section{INTRODUCTION}
\begin{figure*}[!h]
    \centering
    \includegraphics[width=17.5cm]{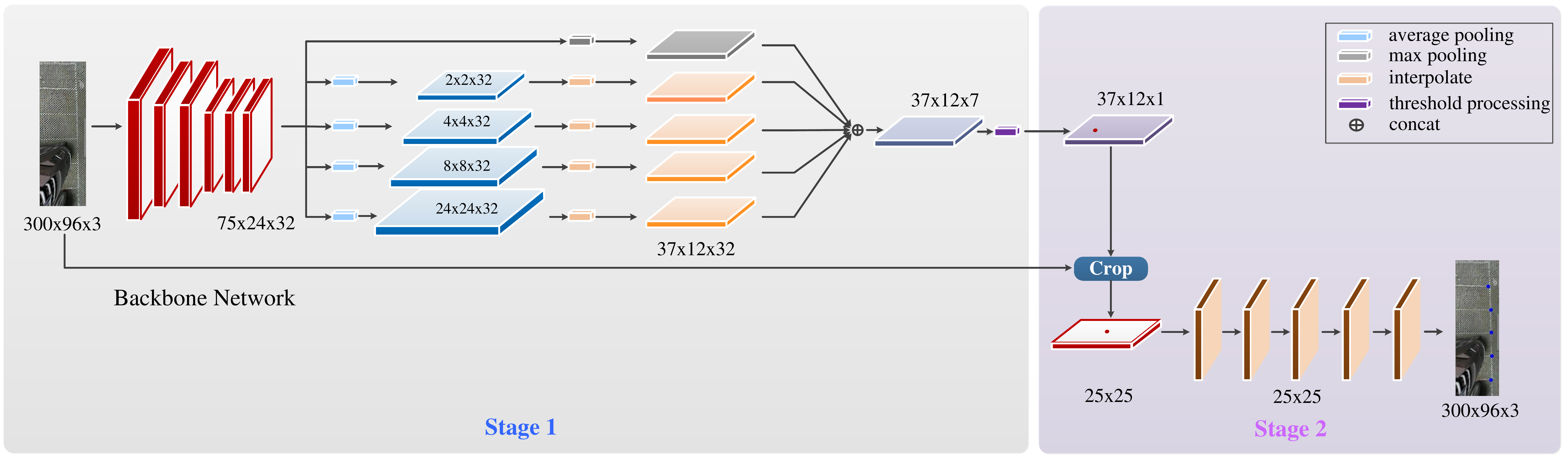}
    \captionsetup{font={small}}
    \caption{The architecture of the presented PSDet. This model is a cascade structure, the first-stage major consists of a backbone network, several down sampling operations and the interpolation process. These interpolated feature maps are concatenated to obtain a feature map containing the initial location of the marking points. In the second stage, the rough positions of marking points obtained in the first-stage are utilized as the centers to clip the sub-images, which are taken as the input of the convolutional neural network. Finally the accurate positions of marking points in the sub-images are detected .}
    \label{fig:Figure01}
\end{figure*}

\noindent
Parking slot detection plays a critical role in valet parking, which necessitates the reliable method -- being efficient and universal in practice~\cite{a1, a2,a3,a4}. Recently, lots of works have been proposed to tackle this problem including free-space-based parking slot detection \cite{a5}, \cite{a6}, \cite{a7} and vision-based method\cite{a8}. 

Free-space-based methods utilize sensors to detect free slots whose neighboring slots are occupied. Despite being simple yet effective in some cases, these methods are limited in realistic cases where neighboring slots are also free. Vision-based methods locate parking slots by recognizing parking slot markings from images such as line segments \cite{a9, a10} and marking points\cite{a4, a17}. In contrast with free-space-based method, it shows the great potential for the universal parking slot detection due to the rich contextual information from images, while overcomes the inability of localizing parking slot without nearby slots being occupied. Therefore, we base our method on vision. 

However, it is still very challenging to detect the parking markings because of immense images complexity. Lots of traditional methods have been proposed -- e.g., Hough Transform for line segment detection. However, the shortcomings of these methods have been widely recognized. Despite being efficient, it still has the poor performance due to the considerable variability of parking markings in practice. Nevertheless, the recent deep methods have endowed parking slot detection with the ability of tackling the variability by increasing the capacity of networks. Besides, these methods aim to detect marking points -- intersection of line segments, to leverage point's simplicity. By doing so, marking points based deep methods have dominated the parking slot detection.  

Seminal works including DeepPS~\cite{a4} and DMPR-PS~\cite{a17} have been proposed to identify marking points for parking slot detection. The main difference of these two works lies in the manner of describing marking points. DeepPs~\cite{a4} utilizes the rectangular descriptor to extract the pattern within the rectangular neighborhood of the parking slot vertexes \cite{a4}. However, the rectangular descriptor is sensitive to the change of direction. Thus, directional descriptors with the T/L templates have been applied in DMPR-PS \cite{a17} to describe the vertex patterns. While this descriptor is more robust to the direction variation, it can only extract the vertex patterns of T/L-shaped parking slots, which is not suitable for describing the complex non-T/L-shaped scenarios such as oblique and trapezoid parking slots. In this work, we aim to improve this limitation.

To do so, we argue that there is no fixed pattern for the various parking slot vertexes, which makes it hard to find a universal way to describe different parking slot vertexes. To address this issue, we proposed a deformable circular descriptor in this paper to enable the network to learn the feature patterns of different types of parking slot vertexes. For different types of parking slot vertexes, the corresponding type of feature mode is used as the descriptor of the parking slot vertexes. Therefore, this descriptor can be compatible with different types of parking slot detection tasks and has better generalization ability. The comparison of different parking slot detection methods is depicted in Table \ref{table01}.

Additionally, the computation overhead of network severely restricts the application of deep learning algorithms in practical engineering application. For example, DeepPS \cite{a4} and DMPR-PS \cite{a17} require powerful GPU to run deep learning algorithm. However, the mass-produced embedded environments merely have CPU, or less powerful GPU. Even though DMPR-PS is designed for the task of embedded system, it is still difficult to process real-time detection without powerful GPU. Given this situation, it is imperative to seek a highly efficient way of slot detection algorithm. 

To this end, we tackle the task in the coarse-to-fine style to reduce the model complexity of the networks. Specifically, our algorithm decomposes the task into two stages as shown in Fig. \ref{fig:Figure01}. In particular, the first stage learns to regress the coarse position of the marking points. This denotes that the optimization of the first stage has the fast convergence due to the simplicity of task. The second stage takes as input the cropped sub-images centered at the predicted coarse position and outputs the finer position to further boost the performance -- offset between coarse position and ground truth. Please note that we use the circular descriptor with different sizes for two stages. Coarse stage (i.e., the first stage) utilizes the larger size of circular descriptor than the fine stage (i.e., the second stage). In this way, our network (i.e., PSDet--Parking slot detection) is capable of being real-time while effective in practice. 

Besides, to validate the performance in real-world applications, we collect and annotate a large-scale benchmark dataset -- Parking Slot Detection Dataset (PSDD) which consists of 7 parking scenarios including brick, grass, oblique, trapezoid, open, rectangular and stereo parking slot. We empirically demonstrate the effectiveness and efficiency of our methods on the PSDD and ps2.0 datasets. The experimental results show that PSDet has much smaller computational complexity than other top-performing methods while achieving the competitive performance. In particular, PSDet achieves state-of-the-art precision rate of \textbf{95.67\%} and recall rate \textbf{98.21\%} on PSDD with \textbf{71 kb} parameters and \textbf{25.3} FPS on CPUs -- being much faster than the top performers. 

In short, we summarize our contributions as follows:
\begin{itemize}
\item We present a novel way of describing marking points for parking slot detection, namely circular descriptor. It is capable of accurately describing a more universal parking slot vertex pattern than prior art, such as directional descriptor~\cite{a17} and rectangular descriptor~\cite{a4}.
\item We propose the two-stage PSDet to realize practical parking slot vertex detection -- being real-time while achieving state-of-the-art precision rate and recall rate.
\item We annotate the large-scale benchmark dataset -- PSDD which includes 7 different scenarios of parking slot. To the best of our knowledge, it is currently the dataset with the largest size and most types of parking slots from real world. We release the dataset for the benefit of community. 
\end{itemize}
\section{RELATED WORK}

\begin{table*}[ht]
\begin{center}
\caption{Comparison of different parking slot detection methods.}
\resizebox{\textwidth}{!}{%
\begin{tabular}{@{}llll@{}}
\toprule
\multicolumn{1}{c}{\multirow{2}{*}{\textbf{Free-space based detection}}} & \multicolumn{3}{l}{Simple to implement} \\ \cmidrule(l){2-4} 
\multicolumn{1}{c}{} & \multicolumn{3}{l}{Sensitive to the surrounding environments} \\ \midrule
\multirow{9}{*}{\textbf{   Vision based detection}} & \multirow{2}{*}{\textbf{Line Based detection}} & \multicolumn{2}{l}{Sensitive to the appearance of lines, and lines are not uniform} \\ \cmidrule(l){3-4} 
 &  & \multicolumn{2}{l}{Weak illumination robustness} \\ \cmidrule(l){2-4} 
 & \multirow{7}{*}{\textbf{Marking point based detection}} & \multirow{2}{*}{\textbf{Direction descriptor}} & Better detection of T-shaped or L-shaped \\ \cmidrule(l){4-4} 
 &  &  & Not suitable for other type parking slot detection such as oblique \\ \cmidrule(l){3-4} 
 &  & \multirow{2}{*}{\textbf{Rectangular descriptor}} & Weak ability to extract common pattern \\ \cmidrule(l){4-4} 
 &  &  & Sensitive to rotation \\ \cmidrule(l){3-4} 
 &  & \multirow{3}{*}{\textbf{Circular descriptor}} & More stable \\ \cmidrule(l){4-4} 
 &  &  & Better ability to extract common pattern \\ \cmidrule(l){4-4} 
 &  &  & Strong rotation robustness \\ \bottomrule
\end{tabular}%
}
\label{table01}
\end{center}

\end{table*}
\noindent
\paragraph{Dataset for Parking Slot Detection} As the largest and completely labeled public benchmark dataset, ps2.0 \cite{a4} has been widely used for parking slot detection. It contains 12165 surround-view images, which are collected from typical indoor and outdoor parking slots under different lighting conditions, and provides the locations of the marking points. We recognize that the release of ps2.0 has greatly promoted the development of the valet parking field, and provided a benchmark to measure the effectiveness of different methods. However, we observe that the most of parking slots in ps2.0 have simply shapes such as T-shaped and L-shaped. It ignores the complex shapes in real-world such as brick, trapezoidal, stereo and oblique parking slots. Thus, networks trained with this dataset will be biased towards simply shape. To overcome this limitation, we annotate the PSDD to further boost the performance in real-world applications.  

\begin{figure}[!h]
\centerline{\includegraphics[width=7cm]{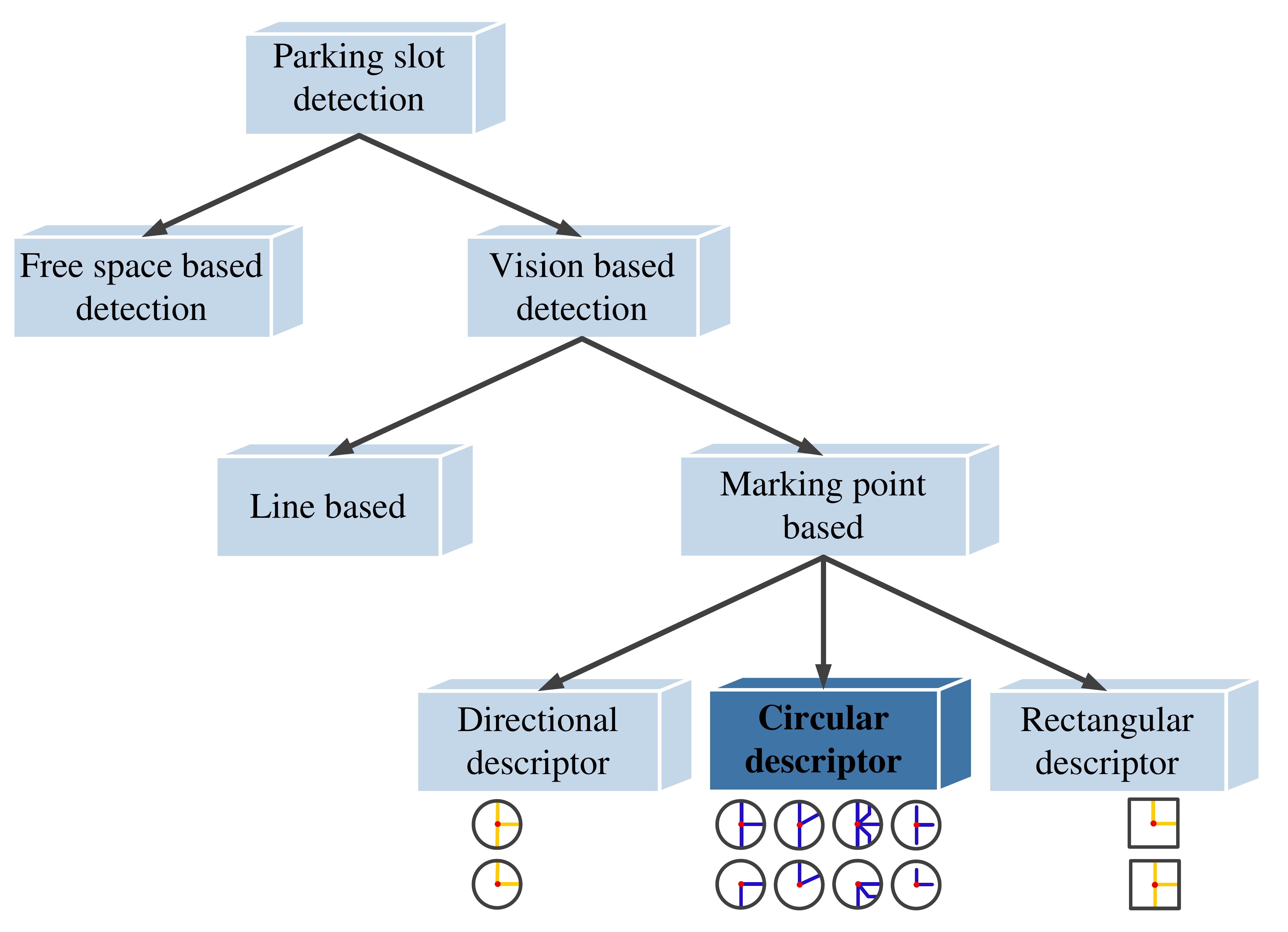}}
\captionsetup{font={small}}
\caption{Different parking slot detection methods.} 
\label{fig:Figure02}
\end{figure}

\paragraph{Descriptor for marking points} The existing parking slot vertex descriptors include rectangular descriptor and directional descriptor \cite{a4}, \cite{a17} -- the details are shown in Fig. \ref{fig:Figure02}. The rectangular descriptor used in DeepPS \cite{a4} is a template for describing features in the rectangular neighborhood of parking slot vertexes, which can extract features of fixed-type parking slots, such as T-shaped, L-shaped, or oblique parking slot. Although it has less limitation than the line-based approach, it is sensitive to the orientation change of the parking line due to its rectangular shape, which degrades generalization ability to different scenes. To address this issue, the directional descriptor is proposed in DMPR-PS \cite{a17}, which is a circular template with T-shape or L-shape vertexes inside. In this method, the position, direction and shape of the parking slot vertex are detected to obtain the parking slots. DMPR-PS has achieved state-of-the-art performance on ps2.0 dataset and argued that the architecture of combining marking point detection and deep learning networks is effective in parking slot detection tasks. Despite the excellent performance of the directional descriptor in T-shaped and L-shaped parking slot detection schemes, it is still limited by the inability of handling complex parking scenarios such as oblique, trapezoid and stereo parking slots.
We demonstrate the comparison of the different detection algorithms and the corresponding supporting scenarios in Table \ref{table02}. 
Inspired by the constraints of the other methods, we propose the deformable circular descriptor to extract the universal vertex feature pattern, which can describe different types of parking slot vertexes. In this way, our method exhibits the capability of detecting parking slots in more complex scenarios.

\begin{table*}[!t]
\centering
\setlength{\abovecaptionskip}{2pt}%
\setlength{\belowcaptionskip}{2pt}%
\captionsetup{font={small}}
\caption{Comparison of the detection algorithms used by different methods and the detection scenarios they can support, where L, R, D and C represent line, rectangular, directional and circular descriptors seperately.}
\begin{tabular}{lcccccccc}
\toprule
Methods & Descriptor & Rectangular & Open & Brick & Grass & Oblique & Trapezoidal & Stereo \\ 
\midrule
Wang et al.'s method~\cite{a10}       & L    &   \checkmark &  & & & & &  \\
Hamada et al.'s method ~\cite{a13}    & L      &\checkmark & & & & &  \\
PSD\-L ~\cite{a21}      & L   & \checkmark & & & & & &  \\
DeepPS ~\cite{a4}      & R     & \checkmark & \checkmark& & & & &  \\
DMPR-PS ~\cite{a17}   & D   & \checkmark &\checkmark &\checkmark & & & &  \\
\textbf{PSDet} & C  & \checkmark & \checkmark & \checkmark & \checkmark & \checkmark & \checkmark & \checkmark \\
\bottomrule
\label{table02}
\end{tabular}
\end{table*}

\paragraph{Efficient Parking Slot Detection} Among the prior parking slot detection methods based on deep learning, DeepPS \cite{a4} is the first network structure to utilize CNN for parking slot detection. In order to improve the efficiency of the parking slot detection task, Lin Zhang et al.  \cite{a17} have proposed a concept named directional marking-point and designed a network structure DMPR-PS. By predicting position, shape and orientation of all directional marking-points of given surround-view image in a single forward evaluation, DMPR-PS obtains the information of marking-points and their neighborhood. As a consequence, DMPR-PS is more efficient than DeepPS. Although DMPR-PS has achieved tremendous progresses in efficiency, it can only perform real-time detection on GPU. However, in the mass production of embedded environment, many platforms only support CPU. Even though some resort to GPU, the computing ability of the GPU is also limited. Some previous researchs \cite{a18}, \cite{a19} have verified that using decoupling method can reduce the difficulty of network convergence by experiments. Inspired by these works, we use the idea of cascade to decouple this problem. In particular, detecting marking point from the input image can be divided into two stages. In the first stage, we employ a circular descriptor to obtain the initial position of the marking point in a $S\times S$ grid. In the second stage, we use the initial position as the center to crop the sub-image in the fixed size from the original image. Then we employ a smaller circular descriptor than the first stage to regress the position shifted by the initial position for a higher accuracy. We corroborated the state-of-the-art effectiveness and efficiency of PSDet in experiments conducted on the benchmark dataset ps2.0 and PSDD.

\section{UNIVERSAL REPRESENTATION OF VERTEX FEATURE }

\noindent
Among the existing parking slot detection approaches, it is difficult to find a universal feature descriptor to describe the parking slot vertexes with complex and variable types. Therefore, we define various types of parking slot vertexes as a universal feature paradigm, and use this paradigm to describe different types of parking slot vertexes.

Compared with the previous rectangular descriptor and directional descriptor, circular descriptors proposed in our paper can describe different types of parking vertex patterns. This chapter will mainly introduce the definition of vertex paradigm and circular descriptors.

\subsection{The Concept of Vertex Paradigm}

\begin{figure}[!ht]
    \centering
    \includegraphics[width=6cm]{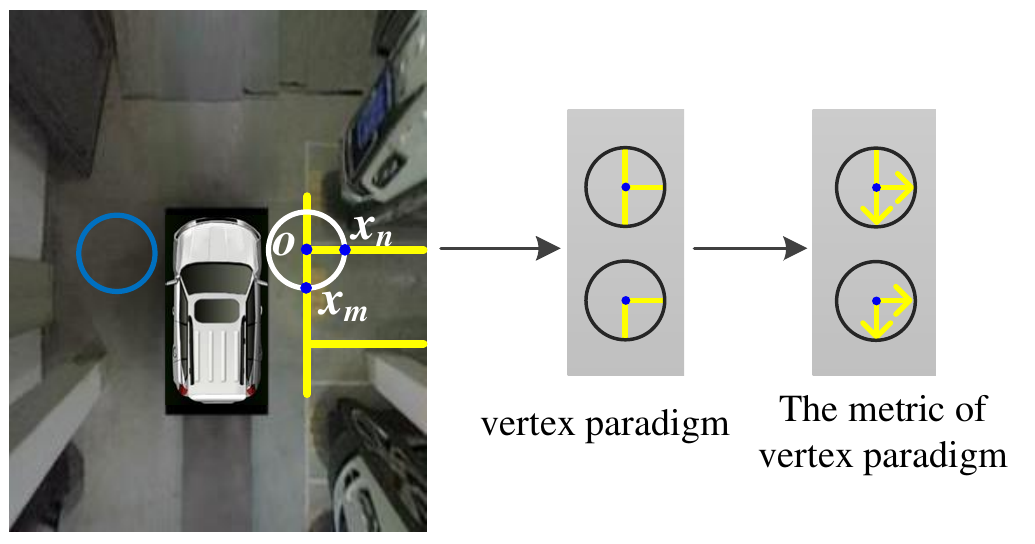}
    \captionsetup{font={small}}
    \caption{The diagram of vertex paradigm. The area contained in the white circle represents $s_p$, and the area contained in the blue circle is an example of $s_{np}$.}
    \label{fig:Figure001}
\end{figure}

\noindent
\textbf{Vertex Paradigm.}\quad  The vertex paradigm is a common pattern of the neighbor pixels around the marking point, which represents the overlapping relationship between the deformable marking-lines around the marking point, as illustrated in Fig. \ref{fig:Figure001}.

\noindent

\textbf{Vertex Area and Non-Vertex Area.}\quad The vertex area is a collection of pixels containing the vertex of the parking line, represented as $s_p$. The Non-Vertex Area is a collection of pixels that are not centered by any parking slot vertex, represented as $s_{np}$, as depicted in Fig. \ref{fig:Figure001}. And $s$ represents the image.

\noindent

\textbf{The Metric of Vertex Paradigm.}\quad The boundary of the vertex area $s_p$ and the marking lines of a parking slot intersect at 2 points  $x_m$ and $x_n$. We take the vertex $o$ as the initial point and the intersection points $x_m$ and $x_n$ as the end points to obtain two vectors $(o,x_m)$ and $(o,x_n)$, respectively, as illustrated in Fig. \ref{fig:Figure001}. The metric of the vertex paradigm is as follows:

\begin{equation}
F=sign(y),
\end {equation}
where: 

\begin{center}
$
sign(y)=
\left\{
             \begin{array}{lr}
             1,  y > 0 &  \\
             0,  y\leq 0 &
             \end{array}
\right.
$
\end{center} 
$y=<(o,x_m),(o,x_n)>$, which represents the inner product between vectors $(o,x_m)$ and $(o,x_n)$. 

\begin{figure}[!h]
    \centering
    \captionsetup{font={small}}
    \includegraphics[width=6.5cm]{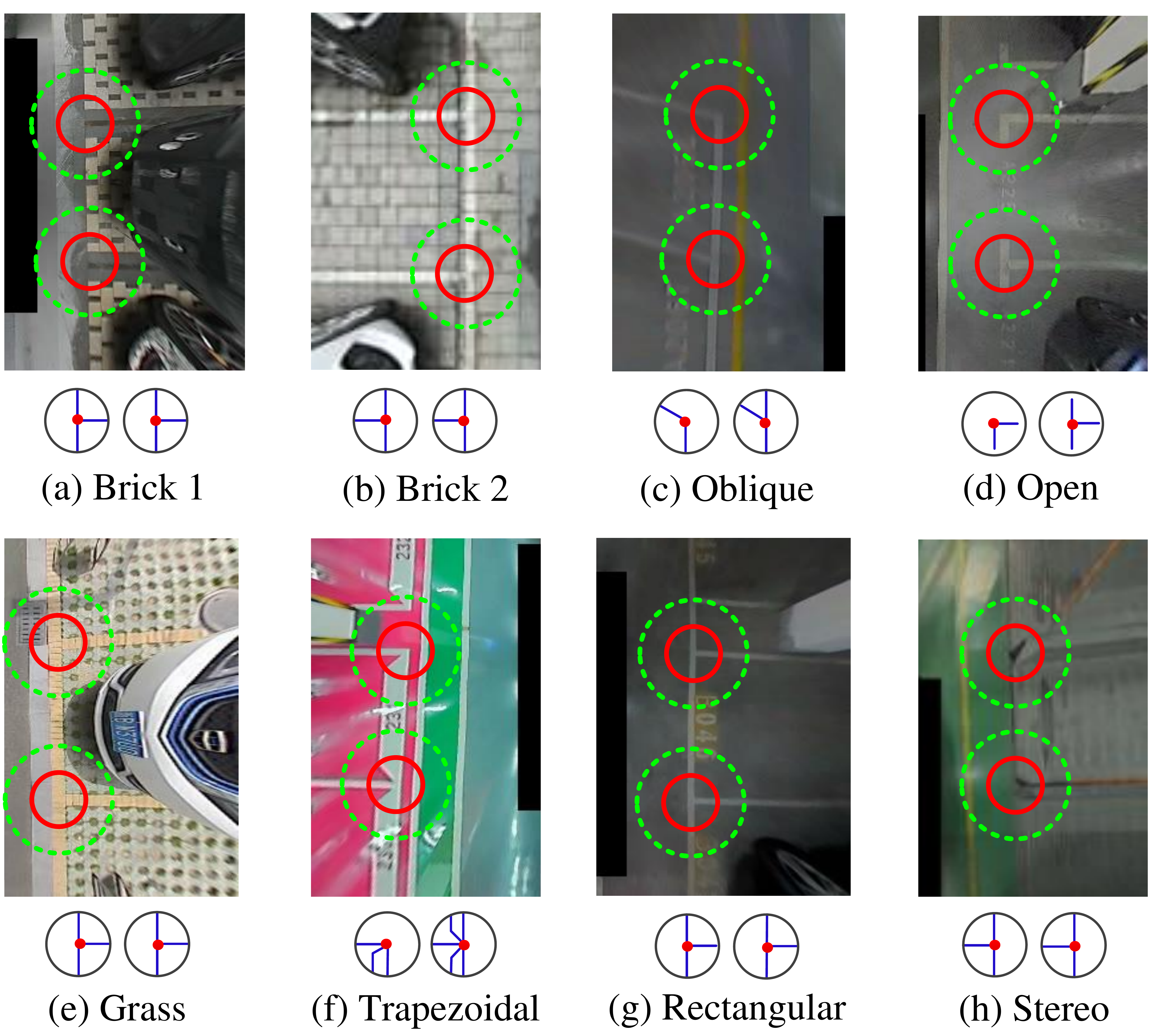}
    \caption{Circular descriptor of different parking slot. The green and red circles on the top of each subgraph represent the circular descriptor used in the first and second stages respectively. The bottom of each subgraph is an abstract representation of the circular descriptors. The blue point is the center of the circle.}
    \label{fig:Figure04}
\end{figure}

\textbf{Circular Descriptor.}\quad To describe the vertex paradigm in the parking slot vertex area $s_p$, we introduce a circular region descriptor. Circular descriptor is a deformable circular template that can contain various types of parking slot vertexes with a sufficiently large radius. Circular descriptors of different parking slots are depicted in Fig. \ref{fig:Figure04}. Different from the T-shaped or L-shaped marking-line pattern around a marking point~\cite{a20}, the circular descriptor is able to extract more common pattern and help to solve the non-L-shaped and non-T-shaped situation, such as oblique, brick and trapezoid etc. Circular descriptors are able to contain various categories of patterns. These circular descriptors can be learned according to the paradigm of corresponding feature patterns given by different labels, as illustrated in Fig. \ref{fig:Figure04}.

\subsection{Characteristics of the Vertex Paradigm}

\noindent
In order to obtain the bounds of the circular descriptor, we will explore the characteristics of the vertex paradigm to determine the boundary range of the circular descriptor.

\begin{figure}[!h]
    \centering
    \includegraphics[width=5cm]{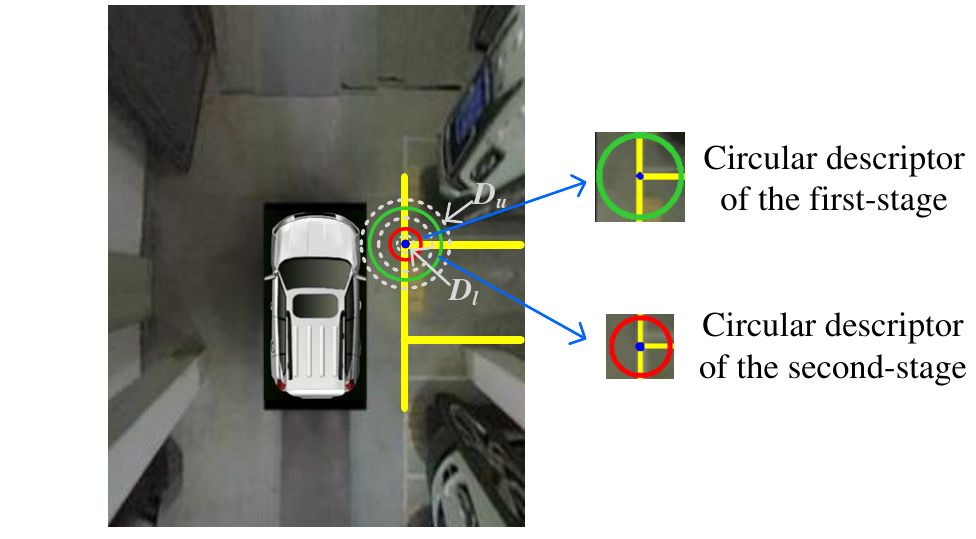}
    \captionsetup{font={small}}
    \caption{The bounds of vertex paradigm. The position of the red circle and green circle are the lower and upper bound respectively. The area bounded by the dotted gray circle is a circular descriptor, and the blue point is the center of the circle.}
    \label{fig:Figure03}
\end{figure}

\textbf{The Lower Bound of Vertex Paradigm.}\quad The lower bound is the minimum area of the vertex paradigm on the image, which is centered at the vertex with the minimum radius that intersects the parking line, as show in Fig. \ref{fig:Figure03}:
\begin{equation}
D_l=argmin(T_l\subseteq C(l)),    
\end{equation}

\noindent
where $T_l$ is the minimum point set containing the vertex, $C(l)$ is the circle with the radius of $l$, $D_l$ is the lower bound of vertex paradigm.

\noindent

\textbf{The Upper Bound of Vertex Paradigm.}\quad The upper bound represents the maximum area of the vertex paradigm on the image, as shown in Fig. \ref{fig:Figure03}:

\begin{equation}
D_u=argmax(T_u\subseteq C(u)),    
\end{equation}

\noindent
where $T_u$ is the maximum point set containing the vertex, $C(u)$ is the circle with the radius of $u$, $D_u$ is the upper bound of vertex paradigm.

\section{PSDet: EFFICIENT AND UNIVERSAL PARKING SLOT DETECTION}

\noindent
In this section, we describe the proposed parking slot vertex detection approach (i.e.,PSDet) in detail.

\subsection{The Structure of PSDet}

\noindent
To achieve an efficient and universal parking slot vertex detection, we propose a two-stage cascade network PSDet. The structure of PSDet is depicted in Fig. \ref{fig:Figure01}.

\textbf{Cascade Network Structure.}\quad It is very challenging to regress the accurate position of marking points from a large image due to immense complexity. In this work, we tackle this problem in the manner of divide-and-conquer. We propose the two-stage PSDet that firstly computes vertex region proposal and then regresses to the accurate vertex location. To more precise, in the first-stage, we extract the approximate region of the vertex appear to coarsely locate the marking points initially. We then crop from the input image sub-images centered around the vertex proposals resulted from the first stage. Additionally, we utilize the second-stage network to regress the accurate vertex position from sub-images in the form of offset to the coarse vertex proposal.

\begin{figure}[!h]
    \centering
    \includegraphics[width=8.5cm]{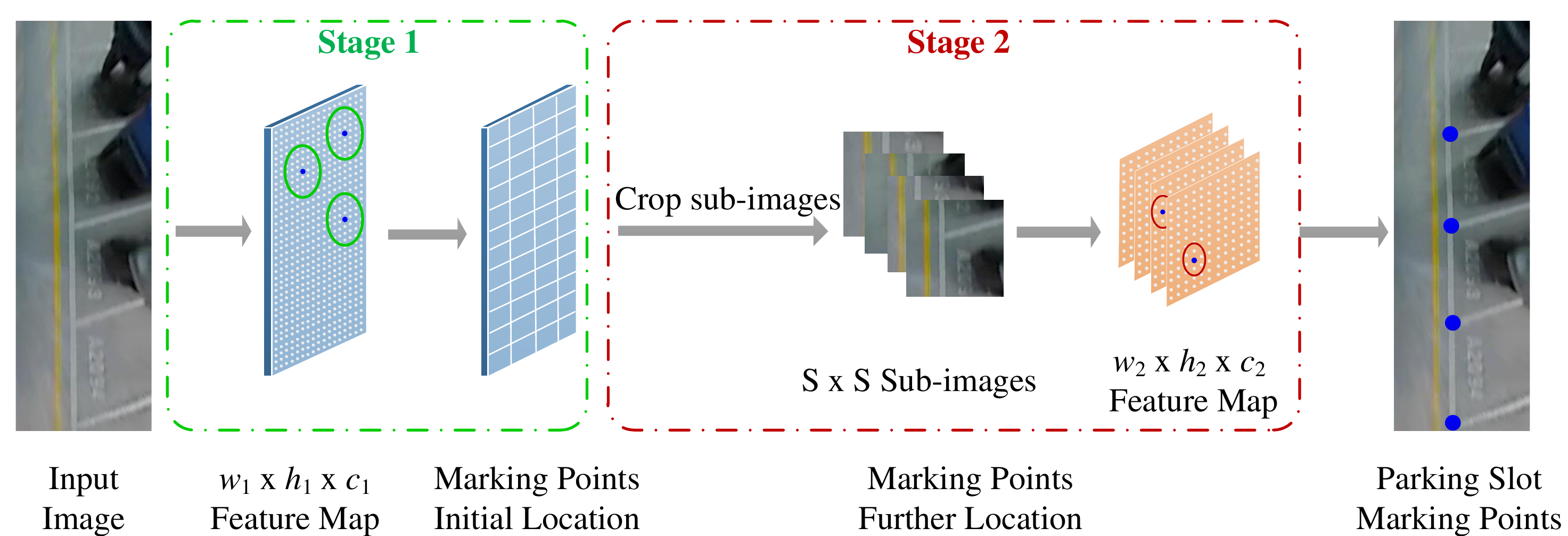}
    \captionsetup{font={small}}
    \caption{The flowchart of of our proposed method PSDet. The green circles and red circles are the circular descriptors used in the first-stage and second-stage respectively.}

    \label{fig:Figure06}
\end{figure}

As shown in Eq. (4), the complexity of detecting marking points directly in the whole image is much higher than detecting marking points in a smaller sub-image. 

\begin{equation}
O(w, h) >> O(w/k_w, h/k_h)+O(w_2, h_2),
\end{equation}

\noindent
where $O(w, h)$ is the complexity of detecting marking points in an original image of size $w \times h$. The size $w \times h$ also indicates the detection range of the first stage. $O(w/k_w, h/k_h)$ is the complexity of detecting marking points in an image of size $w/k_w \times h/k_h$, where $k_w$ and $k_h$ represent the down sampling factor of the original image in the vertical and horizontal direction, respectively. $O(w_2, h_2)$ is the complexity of detecting marking points in an image of size $w_2 \times h_2$, where $w_2 \times h_2$ represents the detection range in the second-stage, as demonstrated in Fig. \ref{fig:Figure06}.
In this way, the proposed PSDet achieves the best performance while being real-time in practice.

\textbf{First Stage.}\quad Given a 320$\times$240 surround-view image $I$, two 320$\times$96 images are cropped with the left and right sides of $I$ as the initial boundaries. Then a set of feature maps is extracted from the 320$\times$96 image, as depicted in Fig. \ref{fig:Figure01}. In addition, the pyramidal network is employed to extract feature maps with different resolutions, which can introduce scaling robustness to the network. Afterwards, these feature maps are interpolated to a fixed size and concatenated into synthesized feature maps. Consequently we obtain a series of feature map of size $w_1 \times h_1 \times c_1$, as shown in Fig. \ref{fig:Figure06}. As an example, We named one of the feature maps as $M$ and the value of point with coordinates $(i,j)$ on $M$ as $M(i,j)$. $M(i,j)$ can be regarded as the point response intensity of input image to the first-stage circular descriptor template. Furthermore, $M(i,j)$ is normalized to [0,1] through softmax, as shown in Eq.(5). Finally, the point position $(i,j)$ whose normalized value $M'(i,j)\geq 0.5$ is retained as vertex proposal of parking slot.

\begin{equation}
M'(i,j)=\frac{e^{M(i,j)}}{\sum_i \sum_j e^{M(i,j)}}
\end{equation}

\noindent

\textbf{Second Stage.}\quad
After obtaining the initial location of the marking points in the first-stage, we utilize the positions of vertex proposals as centers to crop a series of $S \times S$ sub-images from the input image. Then a CNN-based regression model and the second-stage circular descriptor template are used to further detect all vertexes in the sub-images. Finally the position of the point with the highest response intensity on the output feature map is retained as the final position of parking slot vertex, and accordingly correcting the position deviation of the parking slot vertex proposal in the first stage. In this way, the accurate position of the parking slot marking point is detected.

\subsection{Loss}
\noindent
The loss used in PSDet is defined as the sum of squared errors between predictions and ground-truths, and the details will be discussed as follows:

\noindent

\textbf{First Stage Loss.}\quad
The loss used in the first-stage is expressed as following equation:

\begin{equation}
Loss_1=\sum_{i=1}^{w_1}\sum_{j=1}^{h_1}(F_{ij}-\hat{F}_{ij})^2,    
\end{equation}

\noindent
where $w_1$ and $h_1$ represent the width and length of the output feature map from the first-stage. $\hat{F}_{ij}$ is the predicted value in the neighborhood with $(i,j)$ as the center, as depicted by the red circle in Fig. \ref{fig:Figure06}, where $u'$ is the radius of the first-stage circular descriptor. ${F}_{ij}$ is the circular descriptor of the first-stage, as depicted by the green circle in Fig. \ref{fig:Figure03}. We measure the response of each pixel in the output feature map with the method shown in Eq. (1), and obtain the metric result $\hat{F}_{ij}$. With regression objectives defined, the loss function in in the first-stage is defined as the sum of square errors between predictions $\hat{F}_{ij}$ and ground-truths ${F}_{ij}$.

\noindent

\textbf{Second Stage Loss.}\quad
The loss used in the second-stage is defined below:

\begin{equation}
Loss_2=\sum_{i=1}^{w_2}\sum_{j=1}^{h_2}(F_{ij}-\hat{F}_{ij})^2,  
\end{equation}

\noindent
where $w_2$ and $h_2$ represent the width and length of the output feature map from the second-stage, respectively. In this experiment, both width and length are 25. $\hat{F}_{ij}$ is the predicted value in the neighborhood with $(i,j)$ as the center and $l'$ as the radius, as depicted by the red circle in Fig. \ref{fig:Figure06}, where $l'$ is the radius of the second-stage circular descriptor. ${F}_{ij}$ is the circular descriptor template of the second-stage, as illustrated by the red circle in Fig. \ref{fig:Figure03}.

\section{EXPERIMENT}
\subsection{Benchmark Dataset}
\begin{figure}[!ht]
    \centering
    \includegraphics[width=8cm]{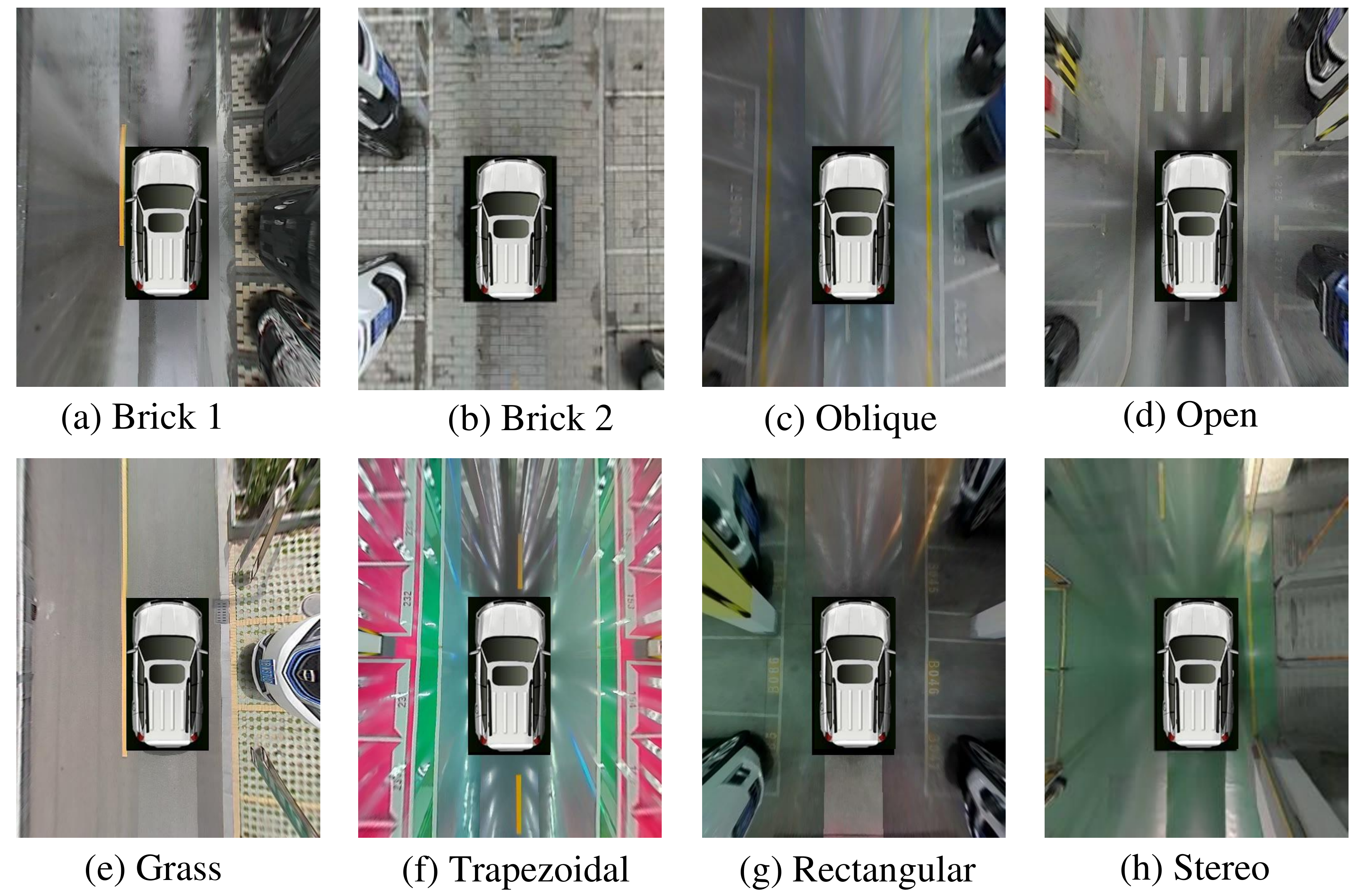}
    \captionsetup{font={small}}
    \caption{A sample set of different types of parking slot in PSDD dataset.}
    \label{fig:Figure08}
\end{figure}

\noindent
In this work, we annotate the large-scale benchmark PSDD which is composed of 14628 calibrated surround-view images collected from typical indoor and outdoor parking slots. We sample the images from 21 video sequences which are captured in 7 different scenarios -- 3 sequences per each scenario. Samples in PSDD are filtered after frame splitting which results in 14628 samples.
The amount of each category of data in the dataset is different, owning to the diverse popularity of the different parking slots in real-world application, such as the rectangular and open parking slots are most frequent. There are 3342 samples in open parking slot category, 5667 samples in rectangular parking slot category, 1242 samples in grass parking slot category, 63 samples in stereo parking slot category, 1946 samples in trapezoidal parking slot category, 500 samples in oblique parking slot category, and 1868 samples in brick parking slot category respectively. A set of samples are shown in Fig. \ref{fig:Figure08}, and the ratio of training set to testing set in all experiment is 1:1. 

\begin{figure}[!ht]
    \centering
    \includegraphics[width=7cm]{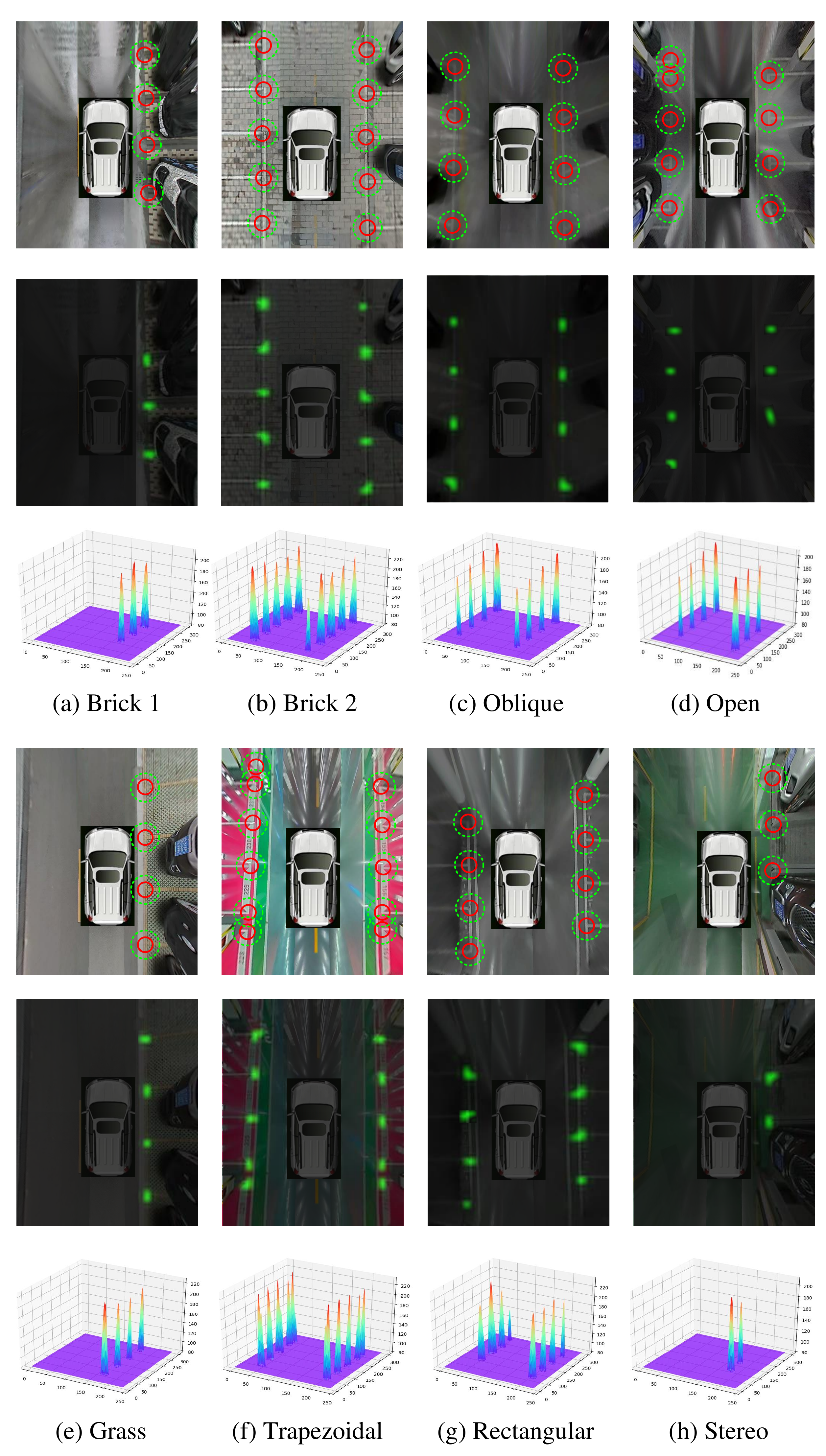}
    \captionsetup{font={small}}
    \caption{Marking point responses of different kinds of parking slots. The first row in each sub-image represents the circular descriptors on the original image, the red circles and green circles represent the circular descriptors of the first stage and second stage respectively. The middle row shows the heatmaps of marking points, and the bottom row shows the response intensity of the marking points. }
    \label{fig:Figure09}
\end{figure}

\subsection{Deformable Marking Point Detection Experiment}
\begin{figure}[!h]
    \centering
    \includegraphics[width=5.5cm]{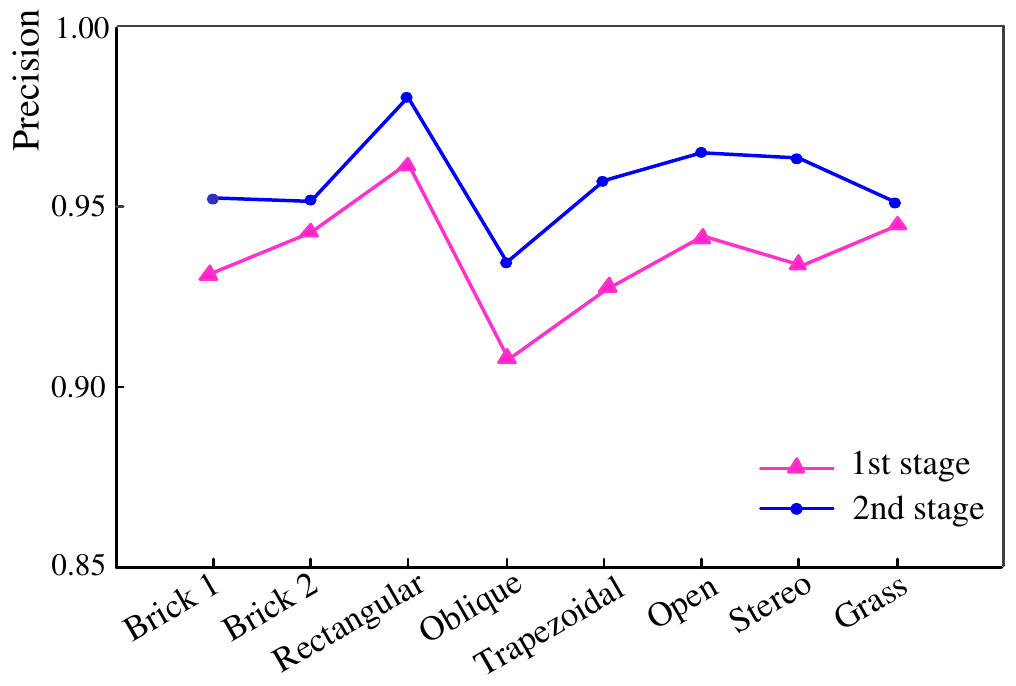}
    \captionsetup{font={small}}
    \caption{Comparison of precision obtained by different stages of PSDet. Pink triangles and blue circles are the precision of first stage and second stage respectively. }
    \label{fig:Figure10}
\end{figure}

\begin{figure}[!h]
    \centering
    \includegraphics[width=5.5cm]{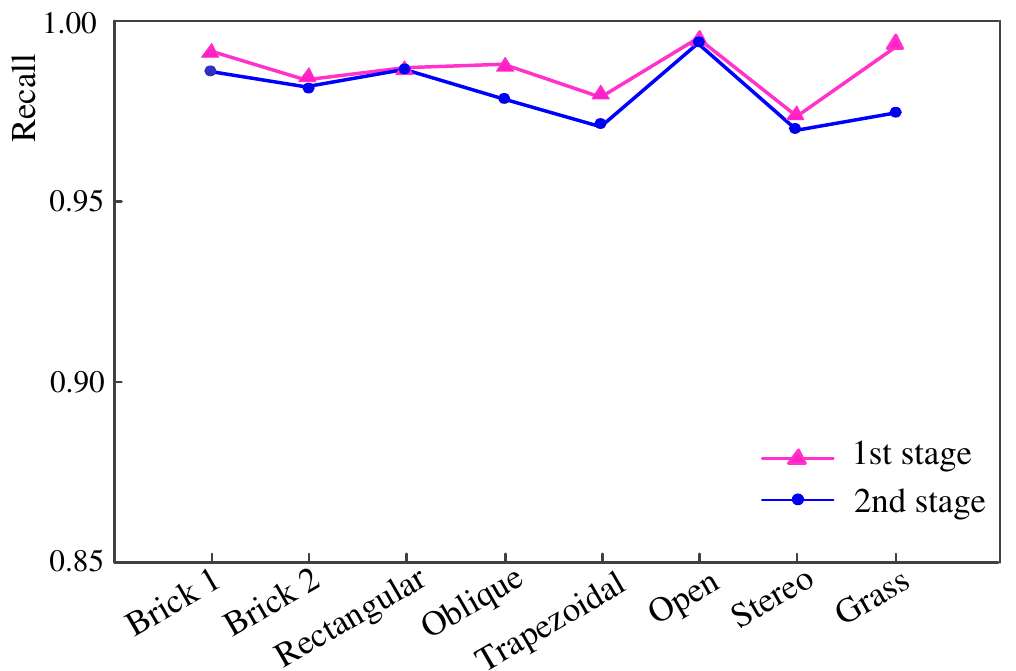}
    \captionsetup{font={small}}
    \caption{Comparison of recall obtained by different stages of PSDet. Pink triangles and blue circles are the recall of first stage and second stage respectively.}
    \label{fig:Figure11}
\end{figure}

\noindent
The circular descriptors and heatmaps of different parking slot are depicted in Fig. \ref{fig:Figure09}. It shows that our approach has better performance. Besides, the precision and recall rate curves obtained after the PSDet first-stage and second-stage are illustrated in Fig. \ref{fig:Figure10} and Fig. \ref{fig:Figure11}, respectively. From all these figures, we observe that the second-stage network significantly improves the precision of the first-stage network while slightly degrades the recall. Please note that the second-stage network is not able to improve the recall due to the hard selection of vertex proposals -- False Positive vertexes are discarded in the second-stage.


\subsection{Parking Slot Detection Experiment}
\noindent
We validate the performance of PSDet on both available benchmarks -- ps2.0~\cite{a22} and PSDD respectively. Precision and recall rates serve as the evaluation metrics. To compare with other methods, we also evaluate the performance of various existing parking slot detection approaches. Table \ref{table02} shows the results on ps2.0. Compared with the most competitive performer -- i.e., DMPR-PS, our PSDet tends to have better recall rate but worse precision. Despite this, we argue that our PSDet achieves the similar precision and recall while with approximately 5$x$ faster speed on simpler dataset. Furthermore, we compare competitors on our more complex PSDD in Table \ref{table03} which has more types of parking slots. In this case, PSDet achieves state-of-the-art in both precision rate and recall rate. 

\begin{table}
\centering
\setlength{\abovecaptionskip}{2pt}%
\setlength{\belowcaptionskip}{2pt}%
\captionsetup{font={small}}
\caption{Performance comparison on ps2.0.}\label{table02}
\begin{tabular}{@{}lccc@{}}
\toprule
Method                   & Descriptor     & Precision & Recall  \\ \midrule
Wang et al.'s method~\cite{a10}      & L     & 98.29\%   & 58.33\% \\
Hamada et al.'s method ~\cite{a13}    & L     & 98.45\%   & 61.37\% \\
PSD\-L ~\cite{a21}          & L     & 98.41\%   & 86.96\% \\
DeepPS ~\cite{a4}          & R           & 98.99\%   & 99.13\% \\
DMPR-PS ~\cite{a17}       & D              & \textbf{99.42\%}   & 99.37\% \\
\textbf{PSDet}            & C           & 98.35\%   & \textbf{99.60\%} \\ \bottomrule
\end{tabular}
\end{table}

\begin{table}
\centering
\setlength{\abovecaptionskip}{2pt}%
\setlength{\belowcaptionskip}{2pt}%
\captionsetup{font={small}}
\caption{Performance comparison on PSDD}\label{table03}
\begin{tabular}{@{}lccc@{}}
\toprule
Method          & Descriptor              & Precision & Recall  \\ \midrule
Wang et al.'s method~\cite{a10}     & L     & 60.13\%   & 10.24\% \\
Hamada et al.'s method ~\cite{a13}  &  L      & 40.96\%   & 12.11\% \\
PSD\-L ~\cite{a21}           &  L          & 65.32\%   & 77.54\% \\
DeepPS ~\cite{a4}            &  R            & 80.23\%   & 78.57\% \\
DMPR-PS ~\cite{a17}         &   D            & 88.45\%   & 81.24\% \\
\textbf{PSDet}              &   C       & \textbf{95.67\%}   & \textbf{98.21\%} \\ \bottomrule
\end{tabular}
\end{table}

\begin{figure}[!h]
    \centering
    \includegraphics[width=7cm]{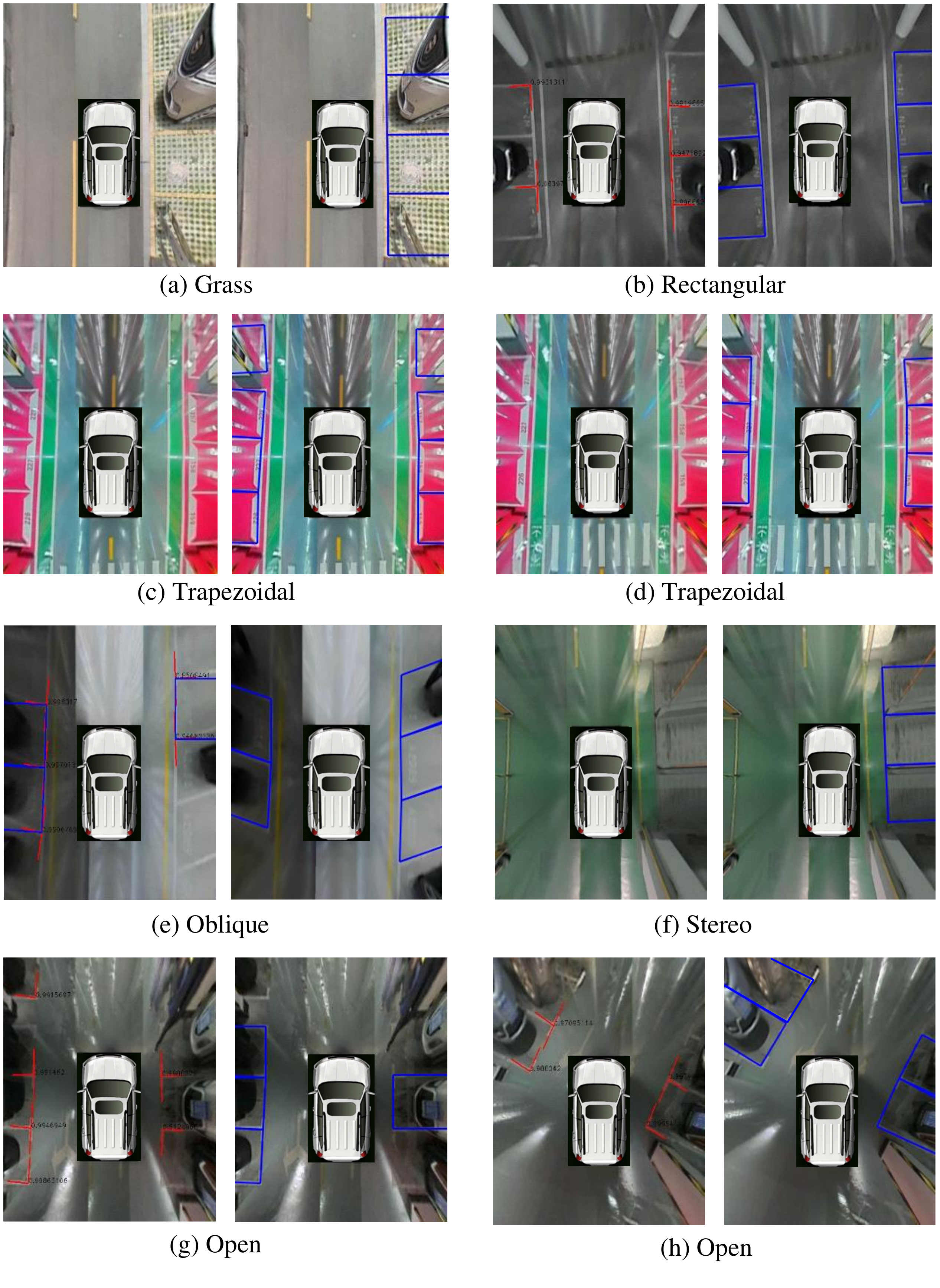}
    \captionsetup{font={small}}
    \caption{Comparison of parking slot detection effect of DMPR-PS (left) and PSDet (right). }
    \label{fig:Figure12}
\end{figure}

We qualitatively compare the performance of DMPR-PS and our PSDet on parking slot detection. As illustrated in Fig. \ref{fig:Figure12}, we observe that PSDet is more accurate and suitble for diverse scenarios. For example, our method is able to accurately detect the parking slots in some extremely challenging cases such as grass scenes with blurry parking lines shown as Fig. \ref{fig:Figure12}(a). Scenes with strong light reflection are shown in Fig. \ref{fig:Figure12}(c) and (d). Reflectional and unclear parking lines of stereo parking slot are shown in Fig. \ref{fig:Figure12}(f).

\subsection{Running Speed of PSDet}
\noindent
PSDet is implemented in C++, and the model is transformed from caffemodel to NCNN format~\cite{a22}. We conduct experiments on the CPU platform Qualcomm 820a, which is low-cost and can be used in mass production. The experimental results show that after converting the model into NCNN format, it can run at speed up to 25.3 FPS on CPU platform Qualcomm 820a (101.8 FPS on GPU platform Qualcomm 820a). In this way, our method is capable of detecting parking slot in real-time. To distinguish our improvement in terms of efficiency, we summarize the time efficiency of other methods on CPU in Table \ref{table04}. All methods are converted to NCNN format before running speed tests. We observe that PSDet achieves the fastest speed. 

\begin{table}[!h]
\centering
\setlength{\abovecaptionskip}{5pt}%
\setlength{\belowcaptionskip}{5pt}%
\captionsetup{font={small}}
\caption{Running speed comparison on CPU}\label{table04}
\begin{tabular}{@{}lcc@{}}
\toprule
Method           & Descriptor       & Speed(FPS)  \\ \midrule
Wang et al.'s method~\cite{a10}   &  L      & 6.1   \\
Hamada et al.'s method ~\cite{a13}  & L       & 7.4   \\
PSD\-L ~\cite{a21}           &    L        & 0.5   \\
DeepPS ~\cite{a4}           &  R             & 1.5   \\
DMPR-PS ~\cite{a17}         &   D            & 5.0   \\
\textbf{PSDet}             &   C        & \textbf{25.3}   \\ \bottomrule
\end{tabular}
\end{table}

\subsection{Potential Practical Application}
\noindent
We show the potential of the proposed method to be applied in practical scenarios. With the known camera parameters, we can easily project the detection results into the world coordinate system. By doing so, we annotate the parking slots in the local semantic map, which is a critical process in self-parking. Furthermore, it is of great significance for the valet parking and autonomous system as well.

\subsection{Failure Cases of PSDet}
\begin{figure}[h]
    \centering
    \includegraphics[width=6cm]{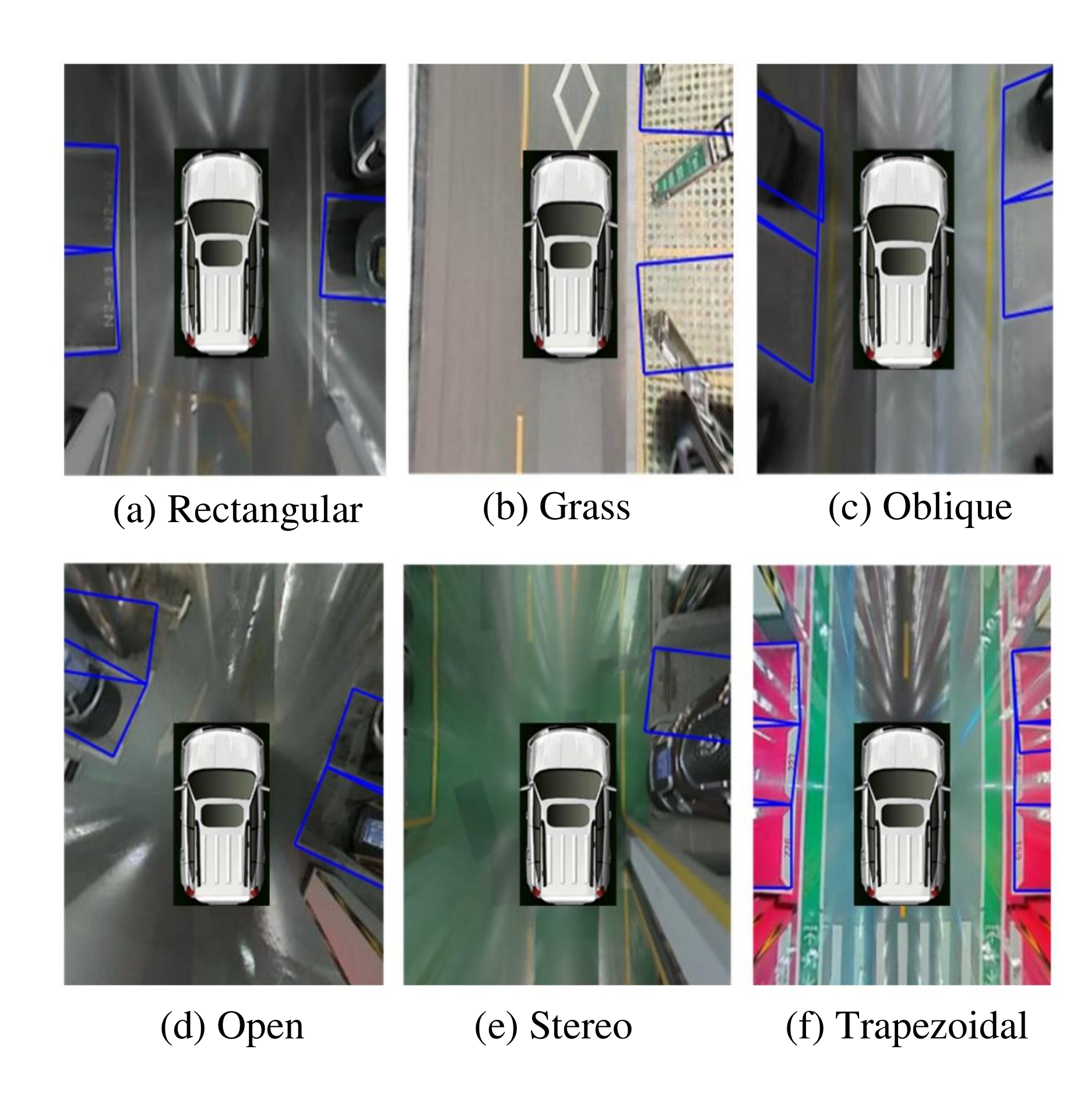}
    \caption{Failure cases detected by PSDet.}
    \label{fig:Figure13}
\end{figure}

\noindent
We summarize two main reasons for failure cases of PSDet: (1), Missing or false detection occurs when image contains adverse factors, such as different light conditions and blur. (2), Severe occlusion of parking lines leads to the inaccurate orientations of them. To name a few examples, we illustrate some typical failure cases in Fig. \ref{fig:Figure13}.

\section{CONCLUSION AND FUTURE WORK}

\noindent
In this paper, we have proposed an efficient and universal method for parking slot detection, named PSDet. 
In contrast with existing work, PSDet is capable of detecting parking slots in various scenarios such as trapezoid, brick and grass parking slots. In addition, PSDet achieves a new state-of-the-art running speed on CPU in embedded platform Qualcomm 820a with a smaller model size -- being much faster than the top performers.
With improved effectiveness and time efficiency, our method therefore provides the technical feasibility for various applications which has limited computing resources such as mobile devices.
Besides, we annotate and release the large-scale benchmark dataset PSDD, which has more types of parking slots and more complex than the existing datasets. We believe that it will be beneficial to the future efforts in enhancing parking slot detection for real-world applications.

\end{document}